\documentclass[conference]{IEEEtran}
\IEEEoverridecommandlockouts
\usepackage{cite}
\usepackage{amsmath,amssymb,amsfonts}
\usepackage{algorithmic}
\usepackage{graphicx}
\usepackage{textcomp}
\usepackage{xcolor}
\usepackage{multirow}
\def\BibTeX{{\rm B\kern-.05em{\sc i\kern-.025em b}\kern-.08em
    T\kern-.1667em\lower.7ex\hbox{E}\kern-.125emX}}
\begin{document}

\title {Churn Prediction via Multimodal Fusion Learning: Integrating Customer Financial Literacy, Voice, and Behavioral Data\\}

\author{\IEEEauthorblockN{1\textsuperscript{st} David Hason Rudd}
\IEEEauthorblockA{\textit{Faculty of Engineering and IT} \\
\textit{The University of Technology Sydney}\\
Sydney, Australia \\ david.hasonrudd@uts.edu.au
}
\and
\IEEEauthorblockN{2\textsuperscript{nd} Huan Huo}
\IEEEauthorblockA{\textit{Faculty of Engineering and IT} \\
\textit{The University of Technology Sydney}\\
Sydney, Australia \\ huan.huo@uts.edu.au
}
\and
\IEEEauthorblockN{3\textsuperscript{rd} Md Rafiqul Islam}
\IEEEauthorblockA{\textit{Information Systems (Data Analytics)} \\
\textit{Australian Institute of Higher Education (AIH)}\\
Sydney, Australia \\ r.islam@aih.edu.au
}
\and
\IEEEauthorblockN{4\textsuperscript{rd} Guandong Xu}
\IEEEauthorblockA{\textit{Faculty of Engineering and IT} \\
\textit{Advanced Analytics institute (AAi)}\\
Sydney, Australia \\ guandong.xu@uts.edu.au
}
}
\maketitle

\begin{abstract}
In today's competitive landscape, businesses grapple with customer retention. Churn prediction models, although beneficial, often lack accuracy due to the reliance on a single data source. The intricate nature of human behavior and high-dimensional customer data further complicate these efforts. To address these concerns, this paper proposes a multimodal fusion learning model for identifying customer churn risk levels in financial service providers.
Our multimodal approach integrates customer sentiments, financial literacy (FL) level, and financial behavioral data, enabling more accurate and bias-free churn prediction models.  
The proposed FL model utilizes a SMOGN-COREG supervised model to gauge customer FL levels from their financial data. The baseline churn model applies an ensemble artificial neural network and oversampling techniques to predict churn propensity in high-dimensional financial data. We also incorporate a speech emotion recognition model employing a pre-trained CNN-VGG16 to recognize customer emotions based on pitch, energy, and tone.   
To integrate these diverse features while retaining unique insights, we introduced late and hybrid fusion techniques that complementary boost coordinated multimodal co-learning. 
Robust metrics were utilized to evaluate the proposed multimodal fusion model and hence the approach’s validity, including mean average precision and macro-averaged F1 score.  
Our novel approach demonstrates a marked improvement in churn prediction, achieving a test accuracy of 91.2\%, a Mean Average Precision (MAP) score of 66, and a Macro-Averaged F1 score of 54 through the proposed hybrid fusion learning technique compared with late fusion and baseline models. Furthermore, the analysis demonstrates a positive correlation between negative emotions, low FL scores, and high-risk customers.

\end{abstract}


\section{Introduction}

Understanding and predicting customer churn has become paramount for long-term success in an increasingly competitive business landscape \cite{b1}. Customer churn, or customer attrition, directly affects a company's revenue and growth potential, making it essential to be accurately predicted and prevented \cite{b2}. 
Furthermore, the human cognitive system is inherently susceptible to cognitive biases, such as anchoring bias, the availability heuristic, and the bandwagon effect~\cite{b9}. These biases can significantly skew customer's feeling and decision-making towards products and services. Misguided customer interactions or suboptimal presentations of complex financial products can trigger these biases, prompting a focus on potential risks over service benefits. For financial service providers, understanding these psychological factors is crucial. It allows them to tailor product design, marketing strategies, and customer relationship management (CRM) to mitigate negative biases and churn rate. Therefore, predicting customer emotion and knowledge about products and services is not just advantageous—it's pivotal for businesses striving for success in today's highly competitive marketplace.

Customer historical data recorded by organizations is increasing exponentially, and data analytics specialists contend that the traditional customer churn analysis method, which heavily relied on transactional data and customer demographics, is no longer a robust solution. A more comprehensive approach is required to better deal with the heterogeneous data input in various modalities.

The current study aims to predict customer churn propensity by incorporating diverse modalities for the first time, to our best knowledge. The proposed approach incorporates survey analysis regarding customer financial literacy level, emotion detection from customer voices (CV), and CRM data to offer a more comprehensive representation of churn risk.

The proposed approach utilizes multimodal machine learning (ML) methods for a more holistic understanding of customer churn. This approach is characterized by considering multiple facets of the customer experience, from interactions with customer service to engagement with a product or service. Multimodality in ML refers to the integration of two or more distinct inputs, recorded in different media formats, within a single ML model. The unique aspect of these inputs is that they cannot be unambiguously mapped into one another using an algorithm~\cite{b11}. We proposed a fusion learning to force similarities of these inputs into a coordinated feature representation space. Subsequently, a decision-level fusion is applied to rank churn risk.

\subsection{Sentiment analysis}

Customer sentiment analysis can be divided into para linguistic, using CV analysis, or linguistic or text based approaches. Recorded customer service calls can be analyzed to understand customer satisfaction. Although sentiment analysis using natural language processing (NLP) can be used to transcribe and analyze these calls; meaning is often also conveyed by tone, pitch, and loudness. Thus, humans provide and receive considerable meta-information during a conversation apart from the text. This is an inherent problem for text-based sentiment analysis in NLP, whereas non-linguistic emotion detection through speech signal processing is somewhat more robust \cite{b3}. Customers negative emotions are often good indicators of product dissatisfaction, and these customers can be considered to have higher churn risk due to their discernible discontent, warranting special attention. To decode emotion from CV, we introduced a voice-based sentiment analysis model\cite{b3}. The central innovation of our work lies in the development of a novel hybrid feature map generator algorithm using harmonic and percussive components of Mel Spectrogram that improves speech emotion recognition (SER). 

\subsection{Financial literacy}

Customer financial literacy can be a crucial factor to predict customer churn within the financial services industry. Financially literate customers are generally better equipped to comprehend the range of products and services available, enabling them to better select the most suitable option(s) to meet their requirements. This reduces dissatisfaction and subsequent churn probability. In contrast, customers with low financial literacy could struggle to understand the options, and hence make sub-optimal decisions regarding financial products and services, leading to dissatisfaction and potential migration to competing financial institutions perceived to offer superior products. We introduced a SMOGN-COREG model to deal with unlabelled examples via semi-supervised regression (SSR) methods~\cite{b8}. We could significantly contribute by enhancing prediction precision beyond relying on labeled data in supervised methods.

\subsection{Multimodal modelling}

The current study focused on analyzing synergistic relationships between predictors in each unimodal context, with the objective is to enhance churn prediction by utilizing a coordinated representation space that forces similarity for three input data formats: finance network, CRM, and inbound audio call data. We investigated the effectiveness for each previously proposed unimodal. Given the identified challenges and opportunities, this study introduces several significant contributions as below: 

\begin{itemize}
    \item [-] In a first-of-its-kind approach, we proposed a multimodal hybrid fusion model that integrates distinct databases, including CRM, CV, and FL data. This enables us to add an essential layer of understanding customer behavior for churn prediction by discerning customer emotion from their vocal attributes and measuring financial literacy level from their account performance and survey data.
    \item [-] We built a specific coordinated feature representation space and translation matrix that reflects data similarity meaningfully to capture interactions between modalities, with the capability of adding future modalities like textual features in the proposed framework.
    \item [-] Identifying significant correlations between negative emotions, low FL, and increased churn risk. 
\end{itemize}

Thus, the proposed multimodal churn model leveraged customer segmentation by detecting negative emotions in CVs during interactions with call center operators, subsequently tailoring the response to address specific customer concerns and requirements, and gaining insights into potential dissatisfaction or frustration causes. 

\section{Recent works}
Considerable research effort is evolving towards more sophisticated churn prediction methods to better draw advantages from data growth. Hence, multimodal churn modelling will continue to advance, providing ever more refined strategies for churn prediction by encompassing various customer experience facets for holistic customer churn understanding~\cite{b14}.

De Caigny explored the benefits from incorporating textual data into customer churn prediction (CCP) models and highlighted superior performance from convolutional neural networks (CNNs) compared with other text mining methods for CCP, achieving 89.87\% accuracy. Outcomes from their study will help guide researchers regarding text mining for improved CCP~\cite{b4}. 

Nhi NY and Liu proposed predicting client churn by analyzing unstructured data, e.g. audio calls, in machine learning models. Their approach expanded traditional methods, which tend to focus on structured data, such as demographics and transactions. In contrast, their proposed model combined text mining with a gradient boosting tree algorithm to improve the client churn model predictive performance. This approach achieved significant predictive accuracy improvement, increasing at least 5\% across various customer datasets~\cite{b5}.

Kimura highlighted benefits from combining boosting algorithms with hybrid resampling methods for CP, which could significantly enhance prediction effectiveness and practicality in both academic and industry settings. The challenge from imbalanced datasets was explicitly discussed, and showed that although hybrid resampling methods, including synthetic minority over-sampling technique (SMOTE), SMOTE-ENN, and SMOTE Tomek-Links, offer significant potential, they are yet to be widely adopted for CCP~\cite{b6}. 
Ahn and Hwang reviewed churn prediction methodologies, highlighting their chronological progression and underscored the necessity for adaptable and flexible approaches contingent on specific data formats. They also showed that very few previous studies adopted a multimodal approach to address churn prediction~\cite{b7}. 

As delineated some recent work in Table~\ref{modalities} from our comprehensive literature review, many studies primarily focus on using singular structured data sources, such as CRM databases, for churn prediction tasks. Conversely, there needs to be more research that employs multiple data sources for the same purpose. To the best of our knowledge, our current investigation is the first that incorporates three discrete input data sets for the churn prediction problem.
\begin{table}[htbp]
\caption{Summary of the comparative studies – Recent Modalities}
\begin{center}
\begin{tabular}{|c|c|c|c|c|}
\hline
\textbf{Ref.} & \textbf{Input} & \textbf{Learning} & \textbf{Prediction} & \textbf{Industry} \\
\hline
\cite{b19} & 1$^{\mathrm{a}}$ & Unimodal & RF & Finance \\
\hline
\cite{b18} & 1 & Ensemble & GBT+k-medios & Telecom \\
\hline
\cite{b17} & 1 & Unimodal & RF & Telecom \\
\hline
\cite{b16} & 1 & Ensemble & DL+LSTM & Game \\
\hline
\cite{b15} & 1 & Ensemble & LSTM+HS & Game \\
\hline
\cite{b5} & 2$^{\mathrm{b}}$ & Feature Fusion$^{\mathrm{d}}$  & GBT & Finance \\
\hline
\cite{b4} & 2 & Feature Fusion & CNN+Logit & Finance \\
\hline
\textbf{Ours} & 3$^{\mathrm{c}}$ & Hybrid Fusion$^{\mathrm{e}}$  & CNNs+DL & Finance \\
\hline
\multicolumn{5}{l}{$^{\mathrm{a}}$1: Structured data (e.g., demographic, account, and CRM data)}\\
\multicolumn{5}{l}{$^{\mathrm{b}}$2: Structured + textual data (e.g., call log script and e-messages)}\\
\multicolumn{5}{l}{$^{\mathrm{c}}$3: Structured + voice + financial literacy (qualitive data)}\\
\multicolumn{5}{l}{$^{\mathrm{d}}$Feature Fusion: Multimodal feature fusion modelling (or early fusion)}\\
\multicolumn{5}{l}{$^{\mathrm{e}}$Hybrid Fusion: Multimodal hybrid (early + late) fusion modelling}
\end{tabular}
\label{modalities}
\end{center}
\end{table}

Therefore, the present study undertook a gap analysis and subsequently proposed an suitable methodology for CCP. This provides significant contributions by incorporating churn predictors, such as customer positive and negative emotional feedback from audio calls, and customer financial literacy level from survey data. The proposed synergistic hybrid feature fusion techniques and subsequent studies building on that framework, will mitigate bias for CCP. 

\section{Methodology}
Our methodology encompasses three modalities to analyze customer behavior: baseline customer churn, SER, and the FL model, which are elaborated in the subsequent sections.  

\subsection{Customer financial literacy modelling}
Considering substantial impact from inadequate FL on customer churn, our first modality emphasizes FL in particular~\cite{b8}. We developed a semi-supervised regression model, (SMOGN-COREG) to measure customer financial literacy derived from a large and unlabeled dataset. The proposed method boosts the learning process by generating synthetic samples for minority class. The SMOGN approach was applied during pre-processing and subsequently combined with non-parametric multi learner semi-supervised regression to enhance model performance, balancing the response variable distribution by increasing the number of rare but important cases in the dataset~\cite{b13}. The model output is FL measurement in a range between 0 and 1. 

\subsection{Emotion recognition modelling}
The second unimodal leverages the voice based sentiment analysis to recognize customer positive and negative emotions during their interaction with the call center~\cite{b3}. The model employs the Mel Spectrogram to provide insight into voice tone, pitch, and tempo; extracting acoustic features and percussive and harmonic components from the signal frequency domain~\cite{b12}.  

We employed a pre-trained CNN to develop a framework decipher emotions from CV, incorporating a feature map generator function to extract harmonic and percussive components via a median filter applied to the signal spectrum's axes. Two feature vectors were subsequently created by averaging these components, and computing the Mel spectrogram logarithm. Finally, we construct a two-dimensional image feature map as input for the pre-trained CNN-VGG16 network to classify emotions into binary outcomes positive and negative classes, represented as 0 and 1, respectively.

\subsection{Baseline churn model}
The third unimodal evaluates customer churn risk using demographic and financial behavior sourced from CRM platforms~\cite{b10}. The proposed approach integrates recursive feature elimination (RFE), SMOTE, and ANN with Deep Learning (ANN DL) to construct the churn classifier. Thus, the proposed framework predicts potential churn within the upcoming six months using data from the prior 12-month window. A customer was deemed a 'churner' if they terminated their account within six-month prediction period, and the outcome is encapsulated in a binary variable, 0 or 1 indicates active or terminated account throughout the subsequent six months.

\subsection{Proposed multimodal fusion learning}
The proposed multimodal is able to analyze customer behavior from a diverse array of data sources highlighting complex interrelationships between different customer behavior facets. The model is designed with different components, including independent unimodal learning, feature representation space, translation or mapping function, and hybrid fusion. 

\begin{figure*}
    \centering
    \includegraphics[scale=0.22] {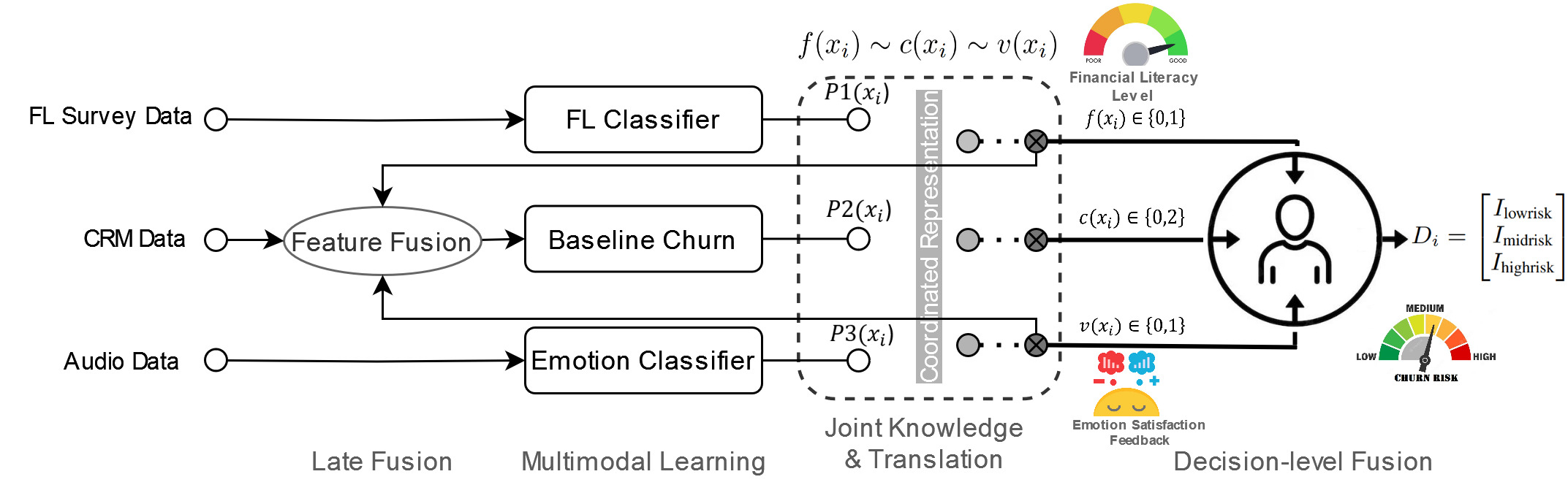}
    \caption{Proposed multimodal hybrid fusion learning method workflow to integrate various modalities}
    \label{fig:Multimodal}
\end{figure*}

Figure~\ref{fig:Multimodal}, shows how gauge customer financial literacy level, customer emotion, and churn propensity were employed to segment customers as low-risk (loyal or non-churner), Mid-risk (potential churner), and high-risk (high probability churner). For example, a high risk churn customer may exhibit low financial literacy, negative sentiment towards the company's offerings, and heightened propensity to churn. 

\subsection*{Hybrid fusion:} The proposed model includes hybrid fusion (multi-level fusion) by incorporating early and late fusion. Hence, hybrid fusion generates input data for the baseline churn model by fusing prediction outcomes from FL and SER models for the CRM dataset.
\subsection*{Late fusion:} We established a prototype multimodal learning model with late fusion to evaluate the proposed hybrid fusion method efficacy compared with other methods. Early fusion data was not utilized as churn model input, with all data sources directly provided to the modalities.

\subsection{Feature representation space and translation}

Extracting specific features associated with each modality presents significant challenges due to inherent disparities in scales, units, and distributions. Therefore, the proposed multi-modal model fusion these heterogeneous features while maintaining their unique insights towards churn modeling. We propose a specific coordinated representation space that reflects data similarity meaningfully to capture interactions between modalities with the ultimate goal to understand complementary customer behavior across various touchpoints, such as call center interaction, FL survey, and historic demographic data. The proposed coordinated representation space can be expressed as 

\begin{equation}
\small
    f(x_{i}) \sim c(x_{i}) \sim v(x_{i})
\label{representation}
\end{equation}

where $x_{i}$ is the modalities and functions $f$, $v$, and $c$ representing independent unimodal learning networks, including FL, SER, and baseline churn model into the coordinated representation space; and $\sim$ is similarity or coordination in the projection into this space.
We employ a translation dictionary to map features extracted from each modality while preserving similarity and meaningful patterns. The mapping matrix is represented through several logical propositions defined by experts, each corresponding to different condition(s) on the unimodal features. Each unimodal output data structure is mapped into two level numeric nominal variables based on a predetermined prediction confidence $P(\text{{pred}})$ threshold, defining a coordinated representation space by logical proposition(s). Constant weights are assigned to each matrix array $C_i$, $F_i$ and $V_i$ depending on the unimodal feature weight in multimodal learning, which can be expressed as, 

{
\small
\[
\forall x_i, P1(x_i) : f(x_i) < P(\text{{pred}}) \implies F_i=1; \neg P1(x_i) \implies 0
\]
\[
\forall x_i, P2(x_i) : c(x_i) \leq P(\text{{pred}}) \implies C_i=0; \neg P2(x_i) \implies 2
\]
\begin{align*}
&\forall x_i, P3(x_i) : v(x_i) \in \{\text{'Happiness'}, \text{'Neutral'}\} \implies V_i=0;&& \\
& \neg P3(x_i): v(x_i) \in \{\text{'Sadness'}, \text{'Anger'}\} \implies V_i=1 && \\
\end{align*}
}
where $P1(x_i)$, $P2(x_i)$, and $P3(x_i)$ denote propositions for each unimodal feature, respectively.
Estimating suitable weights for each feature indicator is challenging. This study assigned weights to the indicators guided by expert knowledge regarding churn modeling and superannuation fund business intricacies. Therefore, implicit feature translation alignment, often incorporated as an intermediate step in mapping data, is imperative for any subsequent task.

\subsection{Hybrid fusion strategy}
The proposed approach employed hybrid fusion to leverage complementary knowledge from unimodal representations while ensuring logical coherence. This incorporates model and decision-level fusion, creating a multi-level fusion framework. We enforced constant fusion weight $C_i + F_i + V_i \geq 0$ and $F_i = V_i$ by assuming fusion weights for FL and SER models had equal influential contribution to enhance baseline model performance. Therefore, a decision fusion $D_i$ was defined to integrate complementary information and hence rank the churn risk as low, mid, or high. The proposed decision fusion matrix $D_i$ can be expressed as

\begin{equation}
\small
D_i = \begin{bmatrix}
I_{\text{{lowrisk}}} \\
I_{\text{{midrisk}}} \\
I_{\text{{highrisk}}}
\end{bmatrix}
= I * \begin{bmatrix}
C_1 + F_1 + V_1 \\
C_2 + F_2 + V_2 \\
\vdots \\
C_i + F_i + V_i
\end{bmatrix}
\label{mm}
\end{equation}

subject to fusion conditions 
\begin{itemize}
\item \textbf{Low risk Churner:}
\begin{equation*}
\small
\begin{split}
I_{\text{{lowrisk}}} = I(D_i = 0) \cdot I(C_i = 0) \cdot I(F_i = 0) \cdot I(V_i = 0) \\
+ I(D_i = 1) \cdot I(C_i = 0) \cdot I(F_i = 1) \cdot I(V_i = 0) \\
+ I(D_i = 1) \cdot I(C_i = 0) \cdot I(F_i = 0) \cdot I(V_i = 1)
\end{split}
\end{equation*}
If $I_{\text{{lowrisk}}} = 1$, customer is classified as low risk.

\item \textbf{Mid risk churner:}
\begin{equation*}
\small
\begin{split}
I_{\text{{midrisk}}} = I(D_i = 2) \cdot I(C_i = 2) \cdot I(F_i = 0) \cdot I(V_i = 0) \\
+ I(D_i = 2) \cdot I(C_i = 0) \cdot I(F_i = 1) \cdot I(V_i = 1) \\
\end{split}
\end{equation*}
If $I_{\text{{midrisk}}} = 1$, customer is classified as mid risk.

\item \textbf{High risk churner:}
\begin{equation*}
\small
\begin{split}
I_{\text{{highrisk}}} = I(D_i = 4) \cdot I(C_i = 2) \cdot I(F_i = 1) \cdot I(V_i = 1) \\
+ I(D_i = 3) \cdot I(C_i = 2) \cdot I(F_i = 1) \cdot I(V_i = 0) \\
+ I(D_i = 3) \cdot I(C_i = 2) \cdot I(F_i = 0) \cdot I(V_i = 1)
\end{split}
\end{equation*}
If $I_{\text{{highrisk}}} = 1$, customer is classified as high risk.
\end{itemize}

Where $D_i \in \{0, 4\}$ is a fusion output weight that embodies ranked risk for each member $i$. $C_i$, represents the fundamental customer churn understanding based on historical patterns; $F_i$ is quantitative financial behavior understanding, i.e., customer FL; $V_i$ is a qualitative measure, reflecting customer emotional state and satisfaction level derived from their phone communication with call center; $I(\cdot)$ is an indicator function where $I(\cdot)=1$ if the condition in the brackets is true and 0 otherwise. 

Thus, the logical operators are designed such that exactly one of $I_{\text{{lowrisk}}}$, $I_{\text{{midrisk}}}$, and $I_{\text{{highrisk}}}$ should = 1 for a given customer risk level query. This ensures each customer is classified into exactly one risk category.

The synergistic integration occurs when these modalities fuse distinct learnings to generate a multimodal co-learning allows a coordinated representation of two modalities FL and SER influence complementary the training of another. Thus, this co-learning strategy mitigates the limitations of relying on a single source of information, thereby enhancing the model's robustness.

\section{Experiments}

\subsection {Data Sources}

We employed real-world data from multiple sources to evaluate the proposed approach efficiency, including customer historic data from finance networks, demographic data from CRM platforms, and voice data, as shown in Table~\ref{database}.

Customer voices were replaced with similar voice from standard emotion databases to maintain privacy while preserving emotional content in the recorded voice files from inbound calls. The proposed method built a CV database leveraging the correlation between negative emotions and high risk churn customers with low financial literacy, as well as the association between positive emotions and low risk churn customers. The EMODB database, a standardized resource for categorizing emotions based on voice recordings, was utilized to label customer emotions~\cite{b14}. The FL database was built by combining financial network data and an online customer survey to measure FL level. After initial data processing, training data was enhanced to (4154, 140). Table~\ref{database} shows the various data sources and their specification.
\begin{table}[htbp]
\caption{Databases Description}
\begin{center}
\begin{tabular}{|c|c|c|c|}
\hline
\textbf{Sub}&\multicolumn{3}{|c|}{\textbf{Specification}} \\
\cline{2-4} 
\textbf{Database} & \textbf{\textit{No. datasets}}& \textbf{\textit{Sources}}& \textbf{\textit{Attributes / Size}} \\
\hline
SDB1& 1 & Financial Networks & 68 /\ $~$64K \\
\hline
SDB2 & 12 & CRM & 87 /\ $~$294m \\
\hline
SDB3 & 1 $^{\mathrm{a}}$ & Audio & 4 /\ $~$110min. voice\\
\hline
\multicolumn{4}{l}{$^{\mathrm{a}}$EMODB database}
\end{tabular}
\label{database}
\end{center}
\end{table}

\subsection{Evaluation criteria}
The mean average precision (MAP) metric was used to objectively evaluate the proposed method solving quality for churn ranking,  
\begin{equation}
\small     
\text{{MAP}} = \frac{1}{Q} \sum_{q=1}^{Q} \frac{1}{m_q} \sum_{k=1}^{n} P(k) \cdot rel(k)
\label{MAP}
\end{equation}

where $Q$ is the total number of queries; $q$ is the current query, $m\_q$ is the relevant churn risk level for the $q$-th query; $P(k)$ is the precision at cut-off $k$ in the list; $rel(k)$ is an indicator function such $rel(k) = 1$ if the churn risk at rank $k$ is a relevant document, and $rel(k) = 0$ otherwise. 
We calculated MAP by taking the mean of average precision (AP) across all risk levels (low, mid, high) for queries or instances. MAP = 1 means the system can find all the churn risk levels accurately. The AP was computed for an individual customer risk rank query as follows.
\begin{itemize}
    \item [-] Calculate the precision for each relevant risk rank retrieved, i.e., the proportion of relevant retrieved risk ranks up to that point, using constant weight for each risk rank. 
    \item [-] Calculate the mean AP across all queries. This provides an overall measure for system performance across a risk rank query set.
\end{itemize}

We also employed the macro-averaged F1 score (MA-F1), since the datasets were typically considerably imbalanced. It was imperative to evaluate smaller class performance with equivalent weighting to larger classes, and F1 score can be calculated independently for each class, 
\begin{equation}
\small
\text{Macro-Averaged F1 Score} = \frac{1}{N} \sum_{i=1}^{N} F1 \text{ Score}_i
\label{Macro_F1}
\end{equation}
where $F1-score_i$ indicates F1 for the $i_{th}$ class and N 
is the number of classes.

\subsection{Correlation analysis}
\begin{figure}[!h]
\centering
\includegraphics [width=\linewidth]{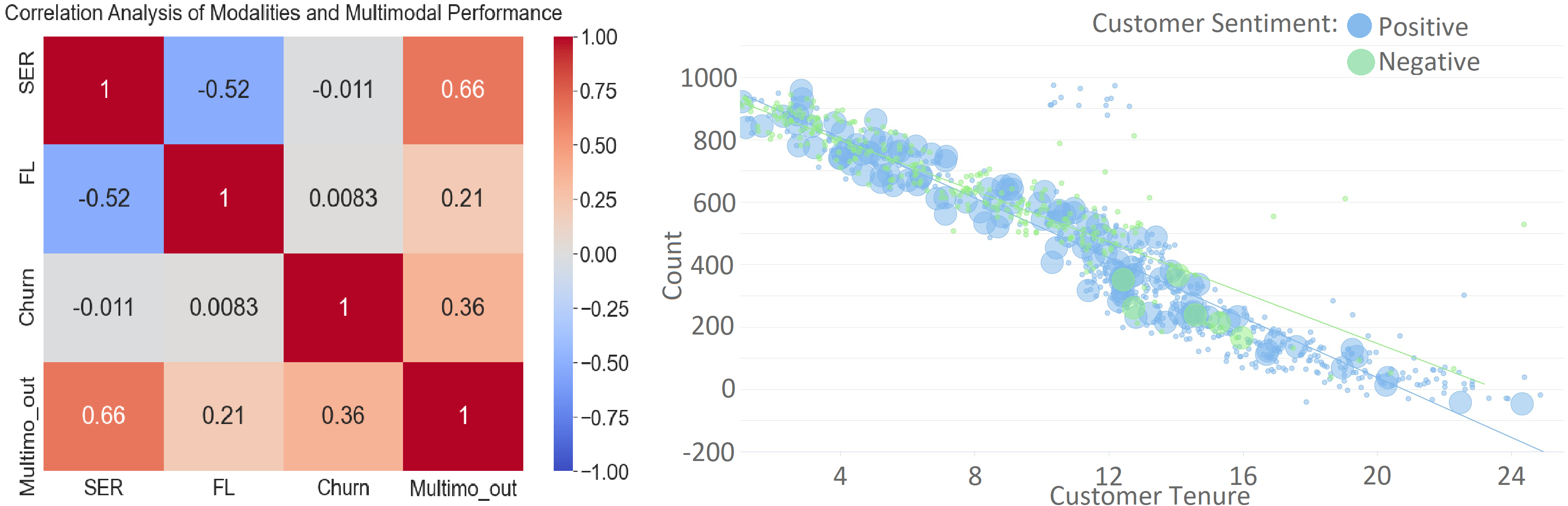}
\caption{a) Strong positive relationship between customer emotion recognition and multimodal learning performance. b) Relation of customer emotion feedback and financial literacy level with customer tenure.}
\label{Multimodal Correlation}
\end{figure}

Figure~\ref{Multimodal Correlation}(a) shows the linear relationship between modalities and the proposed multimodal model output. Thus, the proposed financial literacy and churn unimodal exhibits approximately balanced influences on model performance; whereas the SER unimodal exhibits strong positive correlation = 0.66, confirming significant impact on the multimodal model performance.

Figure~\ref{Multimodal Correlation}(b) shows weightings for low FL and negative emotions (large circles). Thus, clients who renewed their accounts annually exhibited high FL and positive feedback; whereas customers with negative feelings did not renew their accounts or exited.

\subsection{Validation and comparison}
\begin{figure}[!h]
\centering
\includegraphics [width=\linewidth]{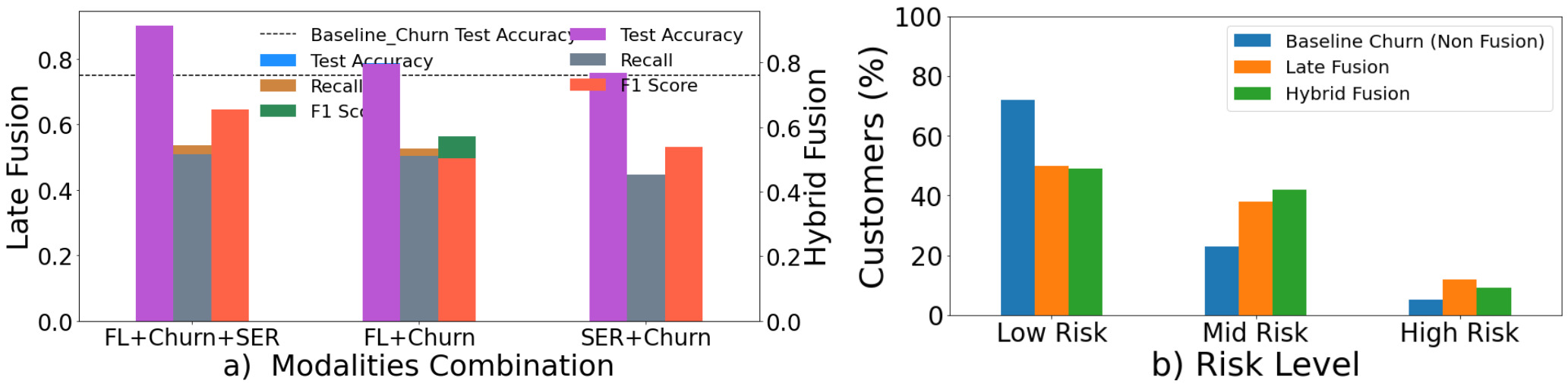}
\caption{(a) Multimodal model performance is significantly improved by combining early and late fusion (i.e., hybrid fusion) considering all features; (b) Risk levels for various multimodal fusion methods, more customers identified as mid and high risk using the proposed hybrid fusion method.}
\label{performance}
\end{figure}

To validate the effectiveness of the proposed method, we investigated the impact of modality integration on performance. Various evaluation metrics were used, including test accuracy, recall, F1-score, AUC, MAP, and MA-F1. The Figure~\ref{performance} (a) shows the highest performance is achieved in hybrid fusion method with contribution of all modalities (FL+Churn+SER) with test accuracy of 91.2\%. 

Moreover, as shown in Figure~\ref{performance}(a) FL has the greatest impact on performance compared with SER. Figure~\ref{performance}(b) shows that the hybrid fusion approach improved churn risk distribution by shifting customers from low into mid and high risk classes. 

\begin{figure}[!h]
\centering
\includegraphics [scale=0.043]{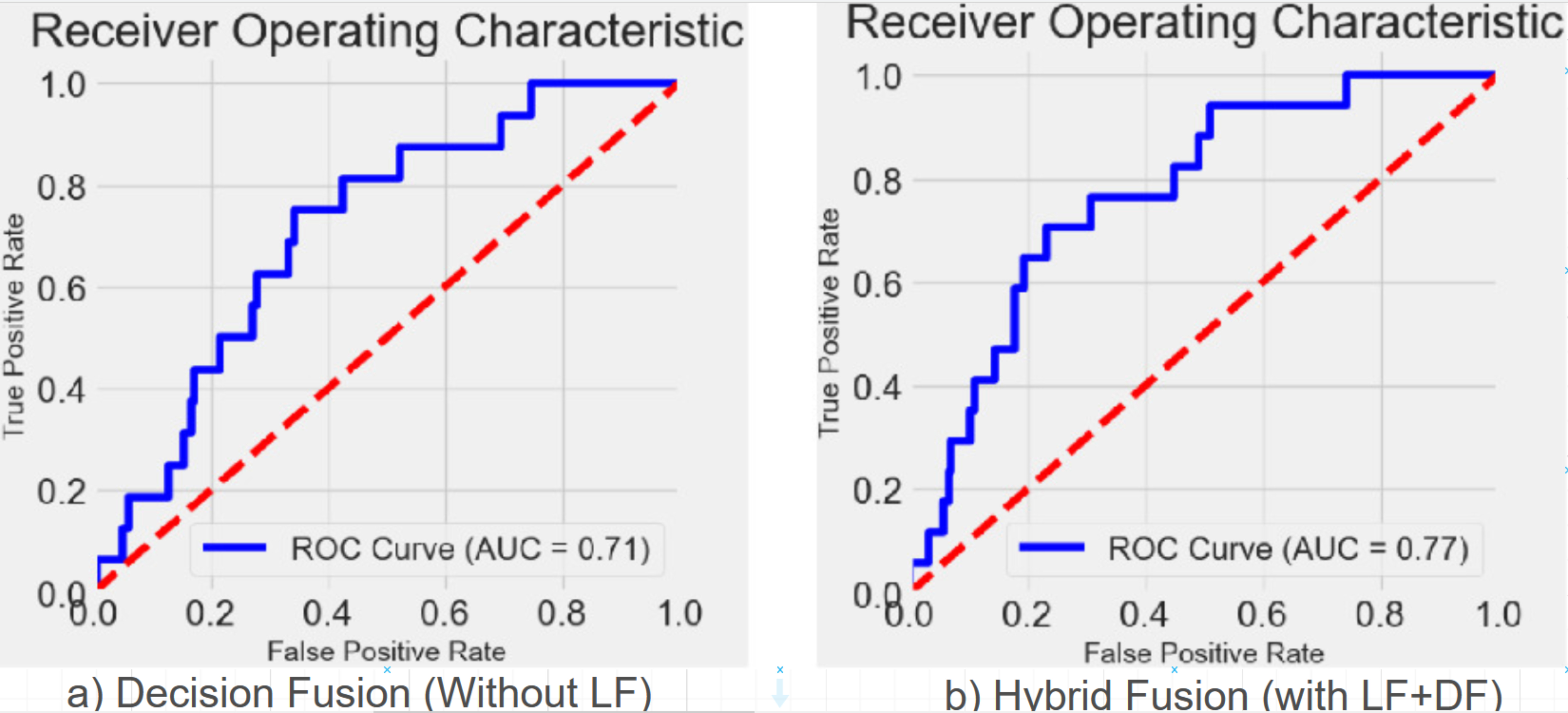}
\caption{Hybrid fusion exhibits higher AUC than other methods considered.}
\label{AUC}
\end{figure}

Figure~\ref{AUC} compares ROC curves for late and hybrid fusion. Higher AUC in Fig.~\ref{AUC} (b) confirms the advantage of integrating multiple modalities, leveraging multi-level fusion benefits.

Table~\ref{Fusion_table} compares risk identification accuracy using MAP and MA F1 metrics. Multimodal learning with hybrid fusion strategy achieved significantly improved results; MAP score = 66 and MA-F1 = 54. 

\begin{table}[htbp]
\caption{Metrics for the different fusion methods considered.}
\begin{center}
\begin{tabular}{|c|c|c|c|}
\hline
\multirow{2}{*}{\textbf{Metrics$^{\mathrm{a}}$}}& \multicolumn{3}{|c|}{\textbf{Fusion Method}} \\
\cline{2-4} 
& \textbf{None} & \textbf{DF (without LF)} & \textbf{HF (LF+DF)} \\
\hline
MAP \% ± STD & 51 ± 0.8 & 65 ± 0.7 & 66 ± 0.1 \\
\hline
MA F1 \% ± STD & 47 ± 0.1 & 47 ± 1.1 & 54 ± 0.6 \\
\hline
\multicolumn{4}{l}{$^{\mathrm{a}}$A higher value implies superior result}
\end{tabular}
\label{Fusion_table}
\end{center}
\end{table}

Python code for our proposed framework with further result visualization is available on our GitHub repository\footnote{https://github.com/DavidHason/multimodal$\_$churn$\_$model}. This resource simplifies the reproduction and enhancement of our study's experimental results. 

\section{Discussion}
This study proposed a multimodal fusion learning model for predicting customer churn. To the authors’ best knowledge, this for the first time an attempt to integrate diverse modalities including customer’s voice (CV), financial literacy (FL) survey, and CRM record data to predict churn risk at three levels (low, mid, and high). These modalities provide distinct insights, thereby contributing to improved churn prediction precision. 

The FL model utilized a SMOGN-COREG supervised model to infer customer FL from their financial behavior data. The baseline churn model, developed on a robust SMOTE and ensemble ANN algorithm combination, accurately predicted churn propensity in the context of massive and high dimensionality data. The proposed SER model employed a pre-trained CNN-VGG16 model to discern customer emotions from their vocal attributes, providing an additional layer to help understand customer behavior. Our findings highlight significant correlations between negative emotions, low FL, and increased churn risk. Furthermore, comparative analysis underscored the superiority of the hybrid fusion technique, achieving a MAP score of 66 and a test accuracy of 91.2\%, thus outperforming both non-fusion and late fusion approaches.

\section{Conclusion and future works}
Despite the promising results, the proposed approach has limitations; in particular, the coordinated representation in the multimodal fusion method could lose valuable intermodal information. Future study will explore a joint representation approach, concatenating features from different modalities before commencing learning, providing the model with a broader, richer data set. This could enhance the learning process efficacy, particularly when employing a powerful model such as Deep ANN.

The second area of potential future work involves extending the multimodal approach by integrating textual features as a fourth modality. It can be represented as $t(x_{i})$ a textual learning unimodal and embedded into existing coordinated representation space which is expressed in equation \eqref{representation}. This would broaden the model's scope and enhance the multifaceted analyses of customer behavior. 

\section*{Acknowledgment}
This work is partially supported by the Australian Research Council under grant number: DP22010371, LE220100078, DP200101374 and LP170100891

\vspace{12pt}
\color{red}

\end{document}